%% file: arxiv_version.tex
\PassOptionsToPackage{table}{xcolor}

\documentclass{article} 
\usepackage[preprint]{colm2025_conference}

\usepackage{arydshln}
\usepackage{xspace}
\usepackage{microtype}
\usepackage{hyperref}
\usepackage{url}
\usepackage{booktabs}
\usepackage{graphicx}
 \usepackage{multirow}
 \usepackage{xcolor}
\usepackage{amsmath}
\usepackage{lineno}
\usepackage{amssymb}
\usepackage{subcaption}
\usepackage{enumitem}
\usepackage{adjustbox}
\usepackage{xltabular}
\usepackage{graphicx}

\definecolor{lightgray}{gray}{0.95}
\definecolor{darkblue}{rgb}{0, 0, 0.5}
\hypersetup{colorlinks=true, citecolor=darkblue, linkcolor=darkblue, urlcolor=darkblue}

\newcommand{\model}{\texttt{\textsc{VLM2Vec-V2}}\xspace}
\newcommand{\benchmark}{\texttt{\textsc{MMEB-V2}}\xspace}
\newcommand{\mmeb}{\texttt{\textsc{MMEB}}\xspace}

\title{VLM2Vec-V2: Advancing Multimodal Embedding for Videos, Images, and Visual Documents}

\author{Rui Meng\textsuperscript{1}\thanks{Contributed equally.} \thanks{Now at Google.} \quad
    Ziyan Jiang\textsuperscript{2}\footnotemark[1] \quad
    Ye Liu\textsuperscript{1} \quad
    Mingyi Su\textsuperscript{3} \quad
    Xinyi Yang\textsuperscript{1} \quad
    Yuepeng Fu\textsuperscript{4} \\
    \textbf{Can Qin\textsuperscript{1} \quad
    Zeyuan Chen\textsuperscript{1} \quad
    Ran Xu\textsuperscript{1} \quad
    Caiming Xiong\textsuperscript{1} \quad
    Yingbo Zhou\textsuperscript{1}} \\ 
    \textbf{Wenhu Chen\textsuperscript{3} \quad
    Semih Yavuz\textsuperscript{1}}
    \\[\medskipamount]
    \textsuperscript{1}Salesforce Research \quad
    \textsuperscript{2}UC Santa Barbara \\
    \textsuperscript{3}University of Waterloo \quad
    \textsuperscript{4}Tsinghua University
}

\newtoggle{comments}
\toggletrue{comments}

\iftoggle{comments}{
    \newcommand{\semih}[1]{\textcolor{blue}{(Semih: #1)}}
    \newcommand{\wenhu}[1]{\textcolor{red}{(Wenhu: #1)}}
}{
    \newcommand{\semih}[1]{}
    \newcommand{\wenhu}[1]{}
}

\begin{document}

\ifcolmsubmission
\linenumbers
\fi

\maketitle

{
\begin{center}
\vspace{-2em}
\url{https://tiger-ai-lab.github.io/VLM2Vec/}
\end{center}
}

\begin{abstract}
Multimodal embedding models have been crucial in enabling various downstream tasks such as semantic similarity, information retrieval, and clustering over different modalities. However, existing multimodal embeddings like VLM2Vec, E5-V, GME are predominantly focused on natural images, with limited support for other visual forms such as videos and visual documents. This restricts their applicability in real-world scenarios, including AI agents, multi-modal search and recommendation, and retrieval-augmented generation (RAG). To close this gap, we propose \model, a unified framework for learning embeddings across diverse visual forms. 
First, we introduce \benchmark, a comprehensive benchmark that extends MMEB with five new task types: visual document retrieval, video retrieval, temporal grounding, video classification and video question answering -- spanning text, image, video, and visual document inputs.
Next, we train \model, a general-purpose embedding model that supports text, image, video, and visual document inputs. Extensive experiments show that \model achieves strong performance not only on the newly introduced video and document retrieval tasks, but also improves over prior baselines on the original image benchmarks. Through extensive evaluation, our study offers insights into the generalizability of various multimodal embedding models and highlights effective strategies for unified embedding learning, laying the groundwork for more scalable and adaptable representation learning in both research and real-world settings.
\end{abstract}

\input{sections/introduction}

\input{sections/benchmark}
\input{sections/method}
\input{sections/experiment}

\input{sections/relatedwork}

\input{sections/conclusion}

\bibliography{colm2025_conference}
\bibliographystyle{colm2025_conference}

\appendix
\input{sections/appendix}

\end{document}

%% file: sections/introduction.tex
\section{Introduction}
Embedding models play a crucial role in connecting data across various modalities. By encoding heterogeneous multimodal data into a shared dense representation space, they enable many down-stream applications like classification, clustering, retrieval, etc. In recent years, we have witnessed significant advances in embedding models, largely driven by the progress of large foundation models. For instance, recent breakthroughs in text embedding \citep{instructor,e5mistral, SFRembedding,LLM2Vec} have been achieved by integrating pretrained large language models with multi-task instruction embedding tuning. Similarly, \citet{vlm2vec,gme, chen-etal-2025-seeing} demonstrated strong performance across multiple text-image tasks by instruction-tuning vision language models (VLMs) into effective embedding models.

Existing multimodal embedding models are trained on datasets like MMEB~\citep{vlm2vec} and M-BEIR~\citep{uniir}, which are focused predominantly on natural images or photographs, sourced from MSCOCO~\citep{mscoco}, Flickr~\citep{flickr30k} and ImageNet~\citep{imagenet} datasets. These datasets fail to cover broader forms of visual information like documents, pdf, websites, videos, slides, etc. The lack of coverage causes the existing embedding models to fall behind on many realistic tasks like article searching, website searching, Youtube video search, etc.


To address these limitations, we introduce \benchmark, an advanced multimodal embedding dataset designed to train and evaluate embedding models across three key visual modalities: images, videos, and visual documents. Expanding on the original MMEB~\citep{vlm2vec} framework, \benchmark broadens the evaluation scope to encompass five new tasks, including four video-based tasks—Video Retrieval, Moment Retrieval, Video Classification, and Video Question Answering — as well as one task centered on visual documents: Visual Document Retrieval. This comprehensive suite of tasks allows for robust assessment of multimodal embedding models across static, temporal, and structured visual data settings.

Built on top of \benchmark, we propose \textbf{\model}, a strong multimodal embedding model fine-tuned from state-of-the-art vision-language models~\citep{qwen2vl}. \model is trained using a mixture of instruction-following tasks spanning multiple task categories, enabling it to produce unified representations for a wide variety of visual modalities. 

Through \model and \benchmark, we aim to investigate the following research questions: How well can a multimodal embedding generalize across diverse visual modalities? What are the key ingredients for training robust and versatile multimodal embedding models? What are the key challenges in representing temporal information in videos and structured information in visual documents?


The contributions of this work are threefold.
\begin{itemize}[
]
\item We propose \benchmark, a comprehensive dataset for systematically evaluating embedding models on diverse tasks involving videos, images, and visual documents.
\item We develop \model, a unified multimodal embedding model that supports diverse input formats, and follows task instructions to produce general-purpose embeddings to support various downstream tasks.
\item Our experiments show that \model outperforms prior baselines across 78 datasets. Through detailed ablations, we identify effective training strategies for learning embedding models across modalities.
\end{itemize}

%% file: sections/benchmark.tex
\section{\benchmark: Expanding Embedding Benchmarks Beyond Image-Text}

\begin{figure}[!ht]
  \centering
    \includegraphics[width=0.95\linewidth]{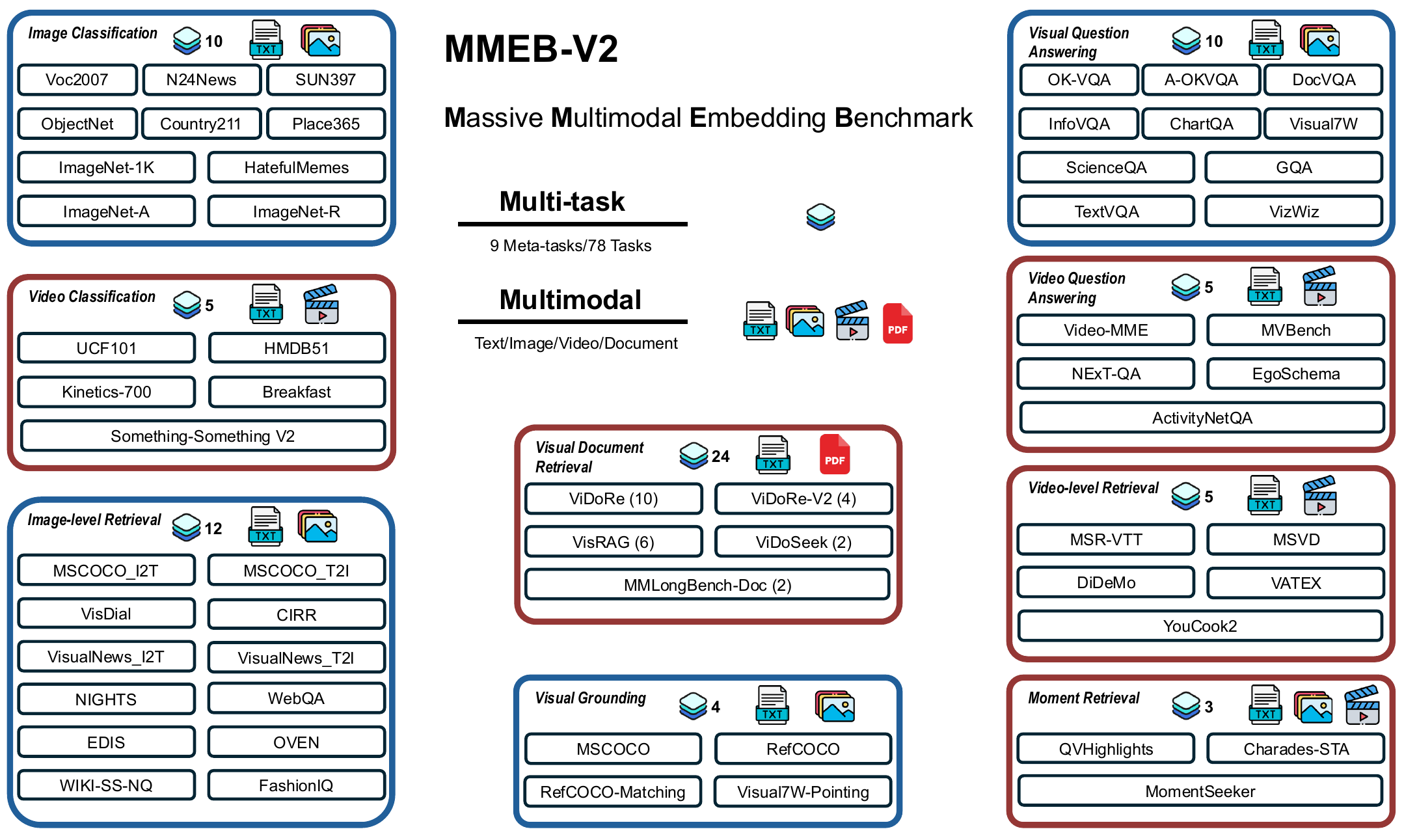}
    \caption{An overview of \benchmark, which includes 9 meta-tasks and 78 tasks in total. In addition to the original \mmeb benchmark, \benchmark introduces five new meta-tasks focused on video and visual documents. Tasks from \mmeb are indicated with blue borders, while newly introduced tasks in \benchmark are marked with red borders.}
    \vspace{-3ex}
    \label{fig:mmeb_pro_overview}
\end{figure}

We introduce \benchmark, a comprehensive dataset designed to evaluate model performance on multimodal embedding tasks involving various combinations of text, image, video, and visual document modalities. In addition to the five task categories that involve natural images and text, \benchmark includes four video understanding tasks and one visual document understanding task. Figure~\ref{fig:mmeb_pro_overview} presents an overview of \benchmark.

Each task presents the model with a query and a corresponding set of candidate responses, where the goal is to select the correct target. The query consists of a combination of an instruction, a textual component, and a video or document. For video-based tasks, videos are represented as sequences of frames sampled at uniform intervals from the raw footage to ensure consistent temporal coverage. Instructions serve to specify the model's objective (e.g., ``Recognize the category of the video contents.''), and query texts can be questions, descriptions, or commands specific to the video (e.g., ``How many red socks are above the fireplace at the end of this video?'', ``Select the clips of videos that contain a dolphin.''). Each task is associated with a specific target type, which varies depending on the nature of the task. For example, in video classification, the model must identify the activity or object class label (e.g., ``Yoga'' or ``Car'').

To make the dataset practical and accessible for future research, we selectively downsample certain datasets to ensure that the full benchmark can be run within a reasonable amount of time. Our datasets span diverse domains, including sports, object recognition, daily-life activities, and movie or TV show clips. These samples are drawn from varied sources such as YouTube, professional productions, and crowd-sourced content, ensuring both diversity and real-world relevance. Summary statistics for each task are presented in Table~\ref{tab:mmeb_v2_stat}. Details about constructing each dataset are provided in Appendix~\ref{sec::appendix_baseline}.

\begin{itemize}[
    itemsep=0pt,        
    parsep=0pt,         
    topsep=0pt,         
    leftmargin=1em      
]
    \item \textbf{Video Retrieval (V-RET)} The query consists of an instruction, a descriptive text related to the video content, and a sequence of video frames. The model must retrieve the correct corresponding video from a pool of thousands of video candidates.
    \item \textbf{Moment Retrieval (M-RET)} The query consists of an instruction, a textual description, and optionally a full video, with the goal of retrieving the temporal segments that best matches the description. The model must select the ground-truth clip from approximately 10 candidate segments within the full video.
    \item \textbf{Video Classification (V-CLS)} Given an instruction and a sequence of video frames, the model is tasked with predicting the correct class label—related to the scene or action depicted—from a set of possible classes.
    \item \textbf{Video QA (V-QA)} The input consists of an instruction, a textual question, and a video. The model must select the correct answer from several options, including one correct choice and many distractors.
    \item \textbf{Visual Document Retrieval (VisDoc)} This task category evaluates the model’s ability to retrieve structured visual documents—such as multi-page PDFs and slide decks—in response to natural language queries. We include five datasets in this benchmark. ViDoRe V1 \& V2~\citep{colpali, vidore_v2} and VisRAG~\citep{visrag} are composed of multiple document QA datasets and cover a broad range of document types and use cases (e.g., charts and slides), though they were not originally designed for retrieval. To complement them, we include ViDoSeek~\citep{vidorag} and MMLongBench-Doc~\citep{mmlongbench}, which provide fine-grained, page-level annotations suitable for retrieval evaluation. Besides, we reformat the two datasets to support both document-level and page-level evaluation. The final VisDoc score is computed as the average of NDCG@5 scores across 24 tasks.
\end{itemize}

\input{tables/mmeb_pro_statistics}

%% file: tables/mmeb_pro_statistics.tex
\begin{table}[!t]
\small
\centering
\resizebox{.95\textwidth}{!}{
\begin{tabular}{lccccc}
\toprule
\textbf{Task} & \textbf{Query MOD} & \textbf{Target MOD} & \textbf{Domain} & \textbf{\#Query} & \textbf{\#Candidates} \\
\midrule
\rowcolor{white} \multicolumn{6}{c}{\textbf{Video Retrieval (5 Tasks)}} \\
\midrule
\rowcolor{lightgray}
DiDeMo          & T     & V       & Open       & 1,004      &       1,004 \\
MSR-VTT         & T     & V       & Open       & 1,000      &       1,000 \\
\rowcolor{lightgray}
MSVD            & T     & V       & Open       & 670        &       670   \\
VATEX           & T     & V       & Open       & 4,468      &       4,468  \\
\rowcolor{lightgray}
YouCook2        & T     & V       & Cooking    & 3,179      &       3,179  \\
\midrule

\rowcolor{white} \multicolumn{6}{c}{\textbf{Moment Retrieval (3 Tasks)}} \\
\midrule
\rowcolor{lightgray}
QVHighlights    & T + V & V       & Vlog/News  & 1,083       &       10 \\
Charades-STA     & T + V & V       & Activity   & 727       &       10 \\
\rowcolor{lightgray}
MomentSeeker    & I + V & V       & Open       & 1,800       &       10 \\
\midrule

\rowcolor{white} \multicolumn{6}{c}{\textbf{Video Classification (5 Tasks)}} \\
\midrule
\rowcolor{lightgray}
Kinetics-700    & V     & T       &   Open     & 1,000       &       700 \\
SSv2            & V     & T       & Human-Object Interaction& 1,000      &       174 \\
\rowcolor{lightgray}
HMDB51          & V     & T       &   Open     & 1,000       &       51 \\
UCF101          & V     & T       &   Open     & 1,000       &       101 \\
\rowcolor{lightgray}
Breakfast       & V     & T       &   Cooking  & 433       &       10 \\
\midrule

\rowcolor{white} \multicolumn{6}{c}{\textbf{Video QA (5 Tasks)}} \\
\midrule
\rowcolor{lightgray}
MVBench             & V + T   & T       &     Spatial/Temporal       & 4,000 & $3 \sim 5$ \\
Video-MME              & V + T   & T       &     Real-world       & 900 & 4 \\
\rowcolor{lightgray}
NExT-QA              & V + T   & T       &    Daily activity        & 8,564 & 5 \\
EgoSchema      & V + T   & T       &     Egocentric       & 500 & 5 \\
\rowcolor{lightgray}
ActivityNetQA      & V + T   & T       &     Activity       & 1000 & 2 \\
\midrule

\rowcolor{white} \multicolumn{6}{c}{\textbf{Visual Document Retrieval (24 Tasks)}} \\
\midrule
\rowcolor{lightgray}
ViDoRe (10)              & T   & D       &   Documents         & 280 - 1,646 & 70 - 999 \\
ViDoRe-V2 (4)              & T   & D       &   Documents         & 52 - 640 & 452 - 1,538 \\
VisRAG (6)   & T   & D       & Documents & 63 - 816 & 500 - 9,590 \\
\rowcolor{lightgray}
ViDoSeek (2)             & T   & D      & Documents & 1,142 & 5,349 \\
MMLongBench-Doc (2)      & T   & D   & Documents & 838 & 6,492 \\
\bottomrule
\end{tabular}
}
\caption{The statistics of \benchmark, which includes \textbf{42} tasks across five meta-task categories in addition to the original \mmeb, are summarized below. Here, we list only the additional datasets introduced beyond those in \mmeb. We consider four modalities (MOD): T (Text), I (Image), V (Video), and D (Visual Document).}
\label{tab:mmeb_v2_stat}
\end{table}
\vspace{-10pt}

%% file: sections/method.tex
\section{Unified Embedding Model for Video, Image, and Visual Document}
Unifying embedding learning across diverse modalities and tasks is inherently challenging due to their distinct structural and semantic characteristics. Our goal is to align data from different modalities in a shared embedding space, while guiding the model's behavior through natural language instructions that define each task.

This section outlines our approach, including multimodal input formatting (Section 3.1), unified encoding with a shared backbone (Section 3.2), instruction-guided contrastive training (Section 3.3), and strategic sampling to balance data sources (Section 3.4).


\subsection{Unified Representation of Multimodal Data}
Our objective is to learn a unified embedding space that supports diverse visual modalities and tasks. This requires a model backbone capable of flexibly encoding interleaved sequences of text, images, and videos, while also handling long-form inputs such as full-length videos and multi-page visual documents. Vision-language models~\citep{llava} have shown strong performance across benchmarks and have proven effective as foundations for multimodal embedding models~\citep{vlm2vec,mmembed}.

Based on these criteria, we adopt Qwen2-VL~\citep{qwen2vl} as the backbone of \model. Qwen2-VL is particularly well-suited for our needs, offering (1) Naive Dynamic Resolution for efficiently processing inputs with variable resolutions, (2) Multimodal Rotary Position Embedding (M-RoPE) to capture spatial and temporal structure, and (3) a unified architecture that integrates 2D and 3D convolutions for consistent image and video understanding. These capabilities enable scalable and generalizable encoding across heterogeneous multimodal data.

\subsection{Contrastive Learning}
We adopt contrastive training to adapt a vision-language model into an embedding model. 

Given a pretrained VLM, we feed query and target into it to obtain the query and target embeddings ($\mathbf{h}_{q_\text{inst}}, \mathbf{h}_{t^+}$)
by taking the last layer vector representation of the last token. To train the embedding model, we adopt the standard InfoNCE loss~\citep{infonce} $\mathcal{L}$ over the in-batch negatives and hard negatives:
\begin{equation} \label{equ:infonce}
    \min\ \  \mathcal{L} = -\log \frac{\phi(\mathbf{h}_{q_\text{inst}}, \mathbf{h}_{t^+})}{\phi(\mathbf{h}_{q_\text{inst}}, \mathbf{h}_{t^+}) + \displaystyle\sum_{t^- \in \mathbb{N}}\phi(\mathbf{h}_{q_\text{inst}}, \mathbf{h}_{t^-})},
\end{equation}
where $\mathbb{N}$ denotes the set of all negatives,
and $\phi(\mathbf{h}_q, \mathbf{h}_t)$ is a function that computes the matching score between query $q$ and target $t$.
In this paper, we adopt the temperature-scaled cosine similarity function as
$\phi(\mathbf{h}_q, \mathbf{h}_t) = \text{exp}(\frac{1}{\tau}\cos(\mathbf{h}_q, \mathbf{h}_t))$,where $\tau$ is a temperature.

\subsection{Multi-modal Data Formatting}

To train a unified embedding model across diverse tasks and modalities, we adopt a standardized format for all query-target pairs. Each training example is denoted as a pair \((q, t^+)\), where \(q\) is the query and \(t^+\) is the positive target. Both components can be a single image, multiple images, text, or interleaved sequences of text and images.

In addition to the raw input pair, we introduce an instruction \(inst\) that specifies the task-specific relationship between the query and target. This guidance helps the model better contextualize and generalize across heterogeneous tasks. We apply the instruction to the original query \(q\) to generate an instruction-conditioned version \(q_{\text{inst}}\):
\begin{equation} \label{equ:instruction_template}
q_{\text{inst}} = \texttt{[VISUAL\_TOKEN]} \ \text{Instruct:}~\textit{\{task\_instruction\}}~\texttt{\textbackslash n}~\text{Query:}~\{q\},
\end{equation}
where \emph{\{task\_instruction\}} is a one-sentence description of the embedding task, e.g. ``'\textit{Find a video that contains the following visual content:}'.

\texttt{[VISUAL\_TOKEN]} is a modality-specific token prepended to indicate whether the visual input is an image or a video. For example, Qwen2-VL uses $<|image\_pad|>$ for image inputs and $<|video\_pad|>$ for video inputs.

Similarly, we optionally apply a simple instruction to the target input \(t^+\) to guide representation learning, e.g.  ``\textit{Understand the content of the provided video:}'':
\begin{equation} \label{equ:target_instruction}
t^+ = \texttt{[VISUAL\_TOKEN]} ~\textit{\{target\_instruction\}}.
\end{equation}

This unified formatting allows the model to handle multimodal inputs consistently while leveraging instruction signals to improve cross-task and cross-modality generalization.




\subsection{Data Sampling Strategies}

To support effective multi-task training over heterogeneous data sources, we design a flexible and scalable data sampling pipeline with two key components.

First, we perform on-the-fly batch mixing guided by a pre-defined sampling weight table. This table specifies the relative probabilities of sampling from each dataset, enabling controlled exposure to different task types. By dynamically drawing samples during training, we ensure balanced coverage and prevent overfitting to any single modality or domain.

Second, we introduce an \textit{interleaved sub-batching} strategy to enhance the hardness and stability of contrastive learning. Specifically, each full batch (e.g., size 1024) is divided into smaller sub-batches (e.g., 8 sub-batches of size 128), where each sub-batch is sampled independently. Compared to per-sample independent sampling, grouping similar samples into sub-batches increases intra-sub-batch homogeneity, which raises the difficulty of contrastive discrimination. At the same time, interleaving multiple such sub-batches preserves cross-task diversity within the full batch, avoiding the instability commonly observed in completely homogeneous batches that originate from a single source. This strategy strikes a balance between sample diversity and structural consistency, fostering more stable and robust optimization dynamics.

%% file: sections/experiment.tex
\section{Experiments}
\subsection{Experiment Setting}
\subsubsection{Training Data}
To train \model effectively across diverse multimodal tasks, we curate a training dataset comprising three main sources: video-language instruction data, visual document retrieval, and image-based vision tasks.

First, we utilize training data from LLaVA-Hound~\citep{llavahound}, which includes synthetic video-caption pairs and video QA examples generated by ChatGPT. Specifically, we use 300k video-caption pairs and 240k video QA pairs. For the caption data, we adopt two formats: using the caption as the query and video as the target in a video retrieval setup, or using the video as the query to retrieve the most relevant textual description from candidate captions.

Second, for visual document retrieval tasks, we incorporate datasets from \texttt{ViDoRe}~\citep{colpali} and \texttt{VisRAG}~\citep{visrag}, including \texttt{colpali train set} (118k), \texttt{VisRAG synthetic} (239k), and \texttt{VisRAG in-domain} (123k), which provide training examples for image-based document understanding and retrieval.

Finally, we include image-text datasets from MMEB-train~\citep{vlm2vec} to support generalization across a wide range of visual understanding tasks, including question answering, classification, retrieval, and visual grounding. These datasets help improve the robustness of the learned embeddings across multiple tasks.

\subsubsection{Training Setting}
We train \model using Qwen2-VL 2B~\citep{qwen2vl} as backbone, a batch size of 1,024 for 2K/5K steps and an interleaved sub-batch size of 64, with the loss temperature set to 0.02. To support scalable training, we use GradCache~\citep{gradcache} to enable large global batch sizes and run all experiments on 8 H100 GPUs. For parameter-efficient training, we apply LoRA tuning with a rank of 16 and scaling factor $\alpha=32$ using the PEFT framework~\citep{peft}. We use 8 uniformly sampled frames to represent each video during both training and evaluation.

We use Hit@1 as the primary evaluation metric for all video and image tasks, measuring the proportion of queries where the correct target is ranked at the top. For visual document tasks, we report NDCG@5 to remain consistent with prior work in this domain.


\subsubsection{Baselines}

We compare against several VLM embedding models, including GME~\citep{gme}, VLM2Vec~\citep{vlm2vec}, and LamRA~\citep{lamra}, most of which are primarily trained on image-text pairs. While these models are not explicitly designed for video tasks, many can be adapted to handle video inputs by encoding multiple frames as sequential images. For video evaluation, GME and LamRA use a single middle frame, while the remaining models use 8 uniformly sampled frames.

In addition, to provide a fair comparison across modalities, we evaluate VLM2Vec-V2 against representative models specialized for each modality. Specifically, we include ColPali~\citep{colpali} (v1.3), a model tailored for document retrieval using a late interaction matching mechanism. 

\subsection{Main Results}    

\input{tables/main_exp}

Table~\ref{tab:main_exp} presents a comprehensive comparison between \model and a diverse set of baseline models across 78 datasets covering image, video, and visual document tasks. Full results are detailed in Appendix~\ref{sec::full_score_table}. \model achieves the highest overall average score (58.0), outperforming multiple strong baselines, including GME, LamRA and VLM2Vec, which were built on the same Qwen2-VL backbone. This highlights the effectiveness of our unified training approach in delivering strong and balanced performance across different modalities and tasks. 
On image tasks, \model shows strong results, outperforming most baselines by a large margin and achieving performance comparable to VLM2Vec-7B despite being only 2B in size. For video tasks, it achieves competitive performance despite being trained on a relatively small amount of video data. In visual document retrieval, \model outperforms all VLM2Vec variants, although still trailing behind ColPali, which is specifically optimized for VisDoc tasks.

\subsection{Ablation Analysis}
\subsubsection{Generalization Across Modalities}
To evaluate the impact of different modality sources on model performance, we consider three types of training data: image-based data (\texttt{Image}), visual document data (\texttt{VisDoc}), and video data (\texttt{Video}). In addition to training models on each individual modality, we construct datasets that combine two modalities (\texttt{Image+VisDoc, Image+Video}) as well as all three modalities (\texttt{Image+VisDoc+Video}). This setup allows us to systematically analyze the contribution of each modality and the effect of multi-modal combinations on model performance. By comparing models trained with single-modality and multi-modality data, we aim to assess how multi-modal training influences generalization and task effectiveness. Vidore-V2 is not used in the ablation studies.

As shown in Table~\ref{tab:modality_comparison}, among single-modality models, training on image data yields the highest average performance. For two-modality combinations, Image+Video slightly outperforms Image+VisDoc, with noticeable gains on image benchmarks in particular. Notably, incorporating all three modalities leads to the best performance on visual document tasks and the highest overall score, highlighting the benefit of comprehensive training data.

\begin{table}[ht]
\centering
\resizebox{0.95\textwidth}{!}{
\begin{tabular}{c|ccc|cc|c}
\toprule
\textbf{Modality} & \textbf{Image} & \textbf{VisDoc} & \textbf{Video} & \textbf{Image+Video} & \textbf{Image+VisDoc} & \textbf{Image+Video+VisDoc} \\
\midrule
Image & 62.5 & 27.9 & 33.9 & \textbf{63.3} & 62.4  & 62.7 \\
VisDoc & 41.5 & 42.6  & 47.9 & 51.9 & 47.4 & \textbf{52.2} \\
Video & 31.5 & 29.1 & 19.9  & 29.7 & \textbf{33.3} & 32.4 \\ \midrule
AVG & 45.2& 33.2 & 33.9& 48.3& 47.7 & \textbf{49.1} \\ 
\bottomrule
\end{tabular}
}
\caption{Performance comparison of models trained on different combinations of modality data.
Rows indicate evaluation performance per modality (Image, Video, VisDoc), while columns represent the modality or modality combinations used during training. }
\label{tab:modality_comparison}
\end{table}

\subsubsection{Ablation Study on Data Sampling Strategies}
As part of our ablation study, we investigate the impact of \textit{interleaved sub-batching} on model performance across the three modalities. When the interleaved sub-batch size (IB) is set to 0, all samples in the batch are randomly drawn from different sources without grouping. A value of 64 indicates that a batch of size 1,024 is divided into sub-batches of size 64, resulting in data from 16 distinct sources per batch. At the other extreme, an IB of 1024 means the entire batch comes from a single source, effectively disabling interleaving. This setup allows us to analyze how different levels of source mixing influence training dynamics and cross-modal generalization.

As shown in Table~\ref{tab:ib_comparison}, increasing the sub-batch size consistently improves performance for both VisDoc and Video. Conversely, the best performance on the Image modality is achieved with a sub-batch size of 64, exhibiting an inverted U-shaped trend—monotonically increasing from 0 to 64, followed by a monotonic decline from 64 to 1024.
\begin{table}[ht]
\centering
\resizebox{0.53\textwidth}{!}{
\begin{tabular}{lccccc}
\toprule
\textbf{Modality} & \textbf{IB0} & \textbf{IB32} & \textbf{IB64} & \textbf{IB128} & \textbf{IB1024} \\
\midrule
Image   & 61.2 & 62.3 & \textbf{63.2} & 62.0 & 60.7 \\
VisDoc & 48.6 & 51.0 & 52.1 & 53.9 & \textbf{54.3} \\
Video  & 34.6 & 33.2 & 33.5 & 34.5 & \textbf{35.4} \\
\bottomrule
\end{tabular}
}
\caption{Performance comparison across different sub-batch size for different modalities.}
\label{tab:ib_comparison}
\end{table}

\subsubsection{Ablation Study on Model Settings}
\begin{figure}[ht]
    \centering
    \begin{subfigure}[t]{0.27\textwidth}
        \centering
        \includegraphics[width=\textwidth]{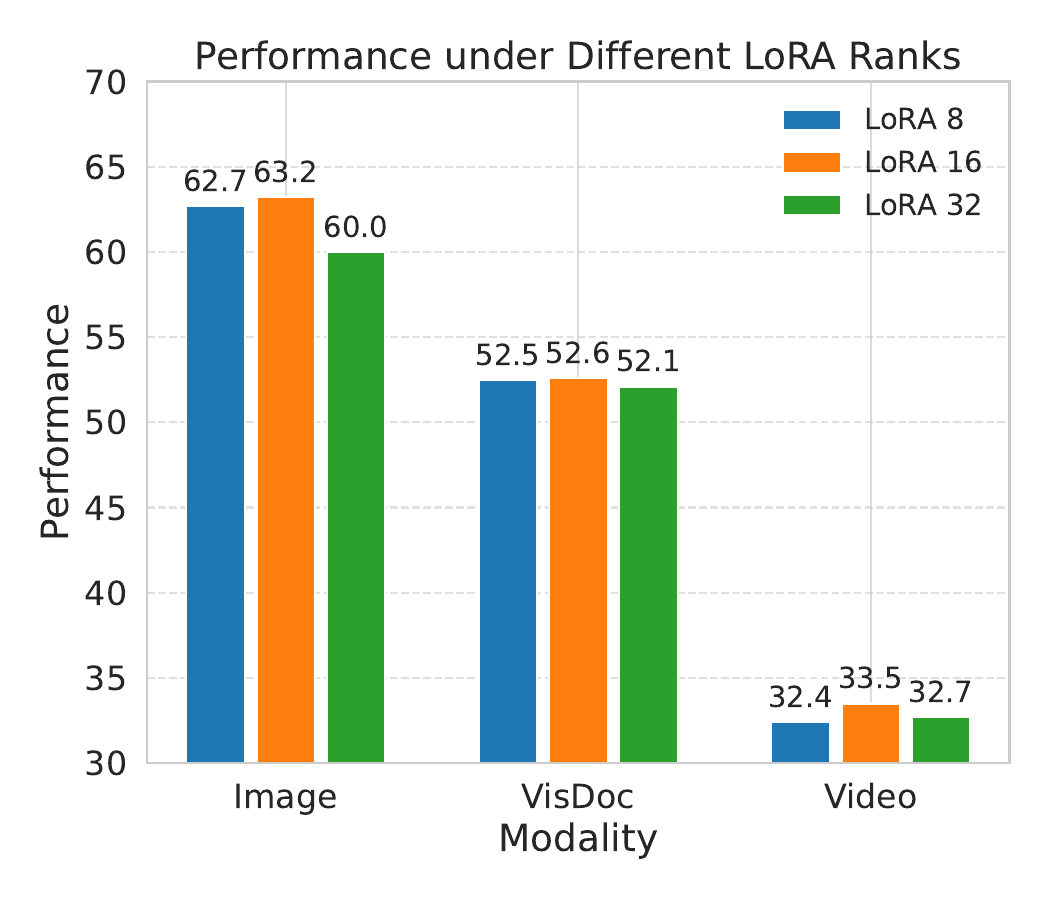}
        \label{fig:lora_ranks}
    \end{subfigure}
    \hfill
    \begin{subfigure}[t]{0.71\textwidth}
        \centering
        \includegraphics[width=\textwidth]{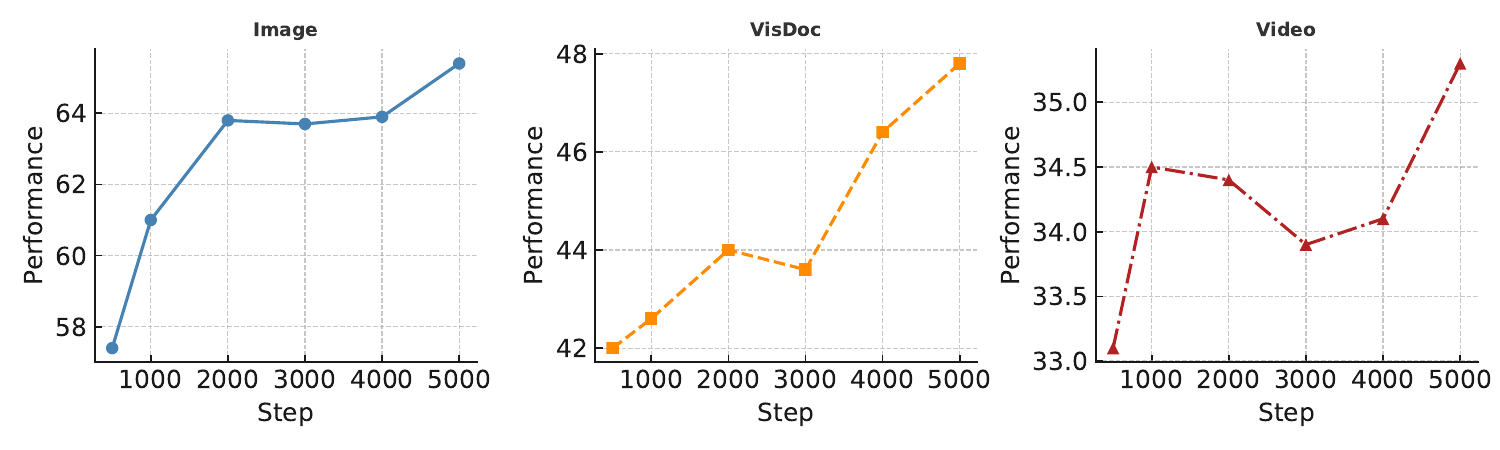}
        \label{fig:training_steps}
    \end{subfigure}

    \caption{The left figure shows performance across LoRA ranks for different modalities, while the right figure illustrates performance trends across training steps.}
    
    \label{fig:multi_modality}
\end{figure}

We investigate the impact of different LoRA ranks (8, 16, 32) on model performance across modalities to understand how the capacity of parameter-efficient tuning affects generalization. As shown in the left part of Figure~\ref{fig:multi_modality}, a LoRA rank of 16 yields the best overall performance across image, video, and visual document tasks. This suggests that a moderate number of tunable parameters is beneficial for handling diverse modalities, while further increasing the rank to 32 does not lead to additional gains.

We also examine performance across training steps to understand how each modality benefits from continued training. As shown in the right part of Figure~\ref{fig:multi_modality}, all three modalities exhibit improved performance with increased training steps. Notably, there is no clear sign of saturation by 5K steps, particularly for VisDoc and Video, suggesting that further gains may be achievable with extended training. We leave a more in-depth exploration of long-horizon training and convergence behavior to future work.



%% file: tables/main_exp.tex
\begin{table}[ht]
\centering
\renewcommand{\arraystretch}{1.2}
\resizebox{\textwidth}{!}{
\begin{tabular}{l ccccc ccccc ccccc c}
\toprule
\multirow{2}{*}{\textbf{Model}} 
& \multicolumn{5}{c}{\textbf{Image}} 
& \multicolumn{5}{c}{\textbf{Video}} 
& \multicolumn{5}{c}{\textbf{VisDoc}}
& \multirow{2}{*}{\textbf{All}}\\
\cmidrule(lr){2-6} \cmidrule(lr){7-11} \cmidrule(lr){12-16}
& \textbf{CLS} & \textbf{QA} & \textbf{RET} & \textbf{GD} & \textbf{Overall} 
& \textbf{CLS} & \textbf{QA} & \textbf{RET} & \textbf{MRET} & \textbf{Overall} 
& \textbf{VDRv1} & \textbf{VDRv2} & \textbf{VR} & \textbf{OOD} & \textbf{Overall} \\
\midrule
\textbf{\# of Datasets} $\rightarrow$ 
& 10 & 10 & 12 & 4 & 36 
& 5 & 5 & 5 & 3 & 18 
& 10 & 4 & 6 & 4 & 24
& 78 
\\
\midrule
\multicolumn{12}{c}{\textbf{Baseline Models}} \\
\midrule
ColPali v1.3 (3B)         & 
40.3 & 11.5 & 48.1 & 40.3 & 34.9 & 
26.7 & 37.8 & 21.6 & 25.5 & 28.2 & 
83.6 & 52.0 & 81.1 & 43.1 & 71.0 & 
44.4
\\
\rowcolor{gray!10}
GME (2B)          & 
54.4 & 29.9 & 66.9 & 55.5 & 51.9 & 
34.9 & 42.0 & 25.6 & 32.4 & 33.9 & 
86.1 & 54.0 & 82.5 & 43.1 & 72.7 & 
54.1      \\
GME (7B)          & 
57.7 & 34.7 & 71.2 & 59.3 & 56.0 & 
37.4 & 50.4 & 28.4 & 38.2 & \textbf{38.6} & 
89.4 & 55.6 & 85.0 & 44.4 & \textbf{75.2} & 
57.8      \\
\rowcolor{gray!10}
LamRA-Qwen2 (7B)        & 
59.2 & 26.5 & 70.0 & 62.7 & 54.1 & 
39.3 & 42.6 & 24.3 & 34.6 & 35.2 & 
22.0 & 11.5 & 37.4 & 21.0 & 23.9 & 
40.4  \\
LamRA-Qwen2.5 (7B)        & 
51.7 & 34.1 & 66.9 & 56.7 & 52.4 & 
32.9 & 42.6 & 23.2 & 37.6 & 33.7 & 
56.3 & 33.3 & 58.2 & 40.1 & 50.2 & 
47.4 \\
\rowcolor{gray!10}
VLM2Vec-Qwen2VL (2B)      & 
58.7 & 49.3 & 65.0 & 72.9 & 59.7 & 
33.4 & 30.5 & 20.6 & 33.0 & 29.0 & 
49.8 & 13.5 & 51.8 & 33.5 & 41.6 & 
47.0    \\
VLM2Vec-Qwen2VL (7B)      & 
62.7 & 56.9 & 69.4 & 82.2 & \textbf{65.5} & 
39.1 & 30.0 & 29.0 & 40.6 & 34.0 & 
56.9 & 9.4 & 59.1 & 38.1 & 46.4 & 
52.3 \\
\midrule
\multicolumn{12}{c}{\textbf{Ours}} \\
\midrule
\textbf{{\model}} (2B) & 
62.9 & 56.3 & 69.5 & 77.3 & 64.9 &
39.3 & 34.3 & 28.8 & 38.5 & 34.9 & 
75.5 & 44.9 & 79.4 & 39.4 & 65.4 & 
\textbf{58.0} \\


\bottomrule
\end{tabular}
}
\caption{Performance comparison between baseline models and our {\model} across image, video, and visual document tasks. CLS: classification, QA: question answering, RET: retrieval, GD: grounding, MRET: moment retrieval, VDR: ViDoRe, VR: VisRAG, OOD: out-of-domain.}
\label{tab:main_exp}
\end{table}

%% file: sections/relatedwork.tex
\section{Related Works}
\subsection{Multimodal Embedding Benchmarks}
Numerous benchmarks have been proposed to evaluate multimodal models, with most early efforts focusing on static image-text pairs. Datasets such as MSCOCO~\citep{mscoco}, Flickr30K~\citep{flickr30k}, and Conceptual Captions~\citep{ccdataset} enabled progress in tasks like image captioning and retrieval. Linear probing is also a common evaluation setting that trains a linear layer for image classification~\citep{clip} to investigate the generalization of representation vectors. More recent benchmarks like M-BEIR~\citep{uniir} and MMEB~\citep{vlm2vec} introduced multi-task evaluations for multimodal embedding models, covering tasks such as retrieval and QA. However, these benchmarks remain limited to static images and short contexts. 

Video-based benchmarks such as MSR-VTT~\citep{msrvtt}, QVHighlights~\citep{qvhighlights}, and ActivityNet Captions~\citep{activitynet} target retrieval and captioning tasks, yet lack unified evaluation frameworks for embeddings. Our \benchmark addresses these gaps by integrating instruction-following tasks across videos and structured documents, providing a comprehensive embedding benchmark for diverse visual modalities.

\subsection{Video Representation Learning}
Video representation learning has evolved significantly, progressing from early convolutional approaches to sophisticated transformer-based architectures. Traditional vision-language models like CLIP~\citep{clip} and BLIP~\citep{blip}, while effective for image-text tasks, often struggle to capture the temporal dynamics inherent in video data. To address this, recent models have been developed to better handle the complexities of video understanding. VideoCLIP~\citep{videoclip}, VideoCoCa~\citep{videococa} integrates contrastive learning with captioning objectives to enhance video-text representation alignment. InternVideo2~\citep{internvideo2} employs a progressive training approach that unifies masked video modeling, cross-modal contrastive learning, and next-token prediction, resulting in superior performance on over 60 video and audio tasks. 

Recent models like LLaVE~\citep{llave} and  LamRA~\citep{lamra}, though trained exclusively on image-text data, have demonstrated the ability to generalize to text-video retrieval tasks in a zero-shot manner. These advancements highlight the ongoing efforts to develop models capable of effectively understanding and representing the complex temporal and semantic information in video data.

\subsection{Visual Document Representation Learning}

Visual document representation learning has become increasingly vital for tasks such as document retrieval, understanding, and retrieval-augmented generation (RAG). Traditional text-based models often struggle to capture the rich visual and structural information present in documents, necessitating approaches that integrate both visual and textual modalities.

One notable advancement is ColPali~\citep{colpali}, which leverages vision-language models to enhance document retrieval efficiency by effectively capturing both textual and visual features. In the realm of retrieval-augmented generation, VisRAG~\citep{visrag} establishes a vision-based RAG pipeline that directly embeds documents as images using vision-language models, thereby preserving the original document information and outperforming traditional text-based RAG systems. Similarly, ViDoRAG~\citep{vidorag} introduces a multi-agent framework tailored for complex reasoning across visual documents, employing a dynamic iterative reasoning process to enhance retrieval and generation tasks. 
Furthermore, benchmarks like MMLongBench-Doc~\citep{mmlongbench} have been developed to assess long-context document understanding with visualizations, providing a comprehensive evaluation framework for multimodal models. 

\subsection{Unified Modality Retrieval}

Unified modality retrieval methods aim to build models capable of retrieving information across multiple data types—such as text, images, audio, and video—within a single framework. Approaches like GME~\citep{gme} and Uni-Retrieval~\citep{jia2025uni} leverage multimodal large language models and prompt-tuning to accommodate diverse queries and modalities, achieving strong performance on universal benchmarks. Meanwhile, methods such as UniversalRAG~\citep{yeo2025universalrag} and UniRAG~\citep{sharifymoghaddam2025unirag} improve retrieval-augmented generation by dynamically routing queries to the most suitable modality and granularity, enhancing both flexibility and accuracy. However, none of these models are designed to unify image, video, and visual document retrieval within a single framework, as our VLM2Vec-V2 does.

%% file: sections/conclusion.tex
\section{Conclusion}

We introduced \benchmark, a comprehensive benchmark for evaluating multimodal embedding models across text, image, video, and visual document modalities. Alongside it, we proposed \model, a strong baseline trained via contrastive learning across a diverse range of tasks and modality combinations. Our extensive experiments demonstrate the effectiveness of \model and the diagnostic value of \benchmark.


%% file: sections/appendix.tex
\newpage
\section{Appendix}

\subsection{Author Contributions}
The VLM2Vec-V2 project was a collaborative effort. Overall project leadership and research guidance were provided by Semih Yavuz, Wenhu Chen, Yingbo Zhou, Caiming Xiong, Ran Xu, and Zeyuan Chen. The specific contributions of the core authors are as follows:

\begin{itemize}
    \item \textbf{Project and Research Leadership:} Rui and Ziyan co-drove the project's technical direction, spearheaded the overall model development and led the creation of the MMEB-v2 benchmark. Semih managed the project's progress, while he and Wenhu provided key research guidance throughout the process.

    \item \textbf{Codebase and Infrastructure:} Rui led the development of the codebase, implementing the v2 refactoring for training and evaluation, and redesigning the data processing infrastructure. Ziyan refactored the evaluation pipeline to integrate the diverse set of tasks. Xinyi contributed to the evaluation for visual document tasks.

    \item \textbf{Benchmark and Data Curation:} The creation of the MMEB-v2 benchmark was a significant team effort.
    \begin{itemize}
        \item \textbf{Video Tasks:} Contributions to the video benchmarks were made by Ziyan (Video Retrieval), Mingyi (Video Classification), Yuepeng (Moment Retrieval), and Rui (Video QA).
        \item \textbf{Visual Documents:} Xinyi curated the majority of the visual document datasets. Ye contributed the ViDoSeek and MMLongBench datasets, and Rui curated ViDoRe-V2. Ziyan developed the corresponding data parsers and evaluation logic.
    \end{itemize}

    \item \textbf{Modeling and Experiments:} Ye conducted the training experiments for most models. Can ran the overall evaluations and reported the scores. Xinyi conducted the evaluation for visual document tasks. The collection of baseline results was a joint effort by Ziyan, Rui, Mingyi, Xinyi, and Yuepeng.

    \item \textbf{Maintenance:} Our team is committed to the long-term and active maintenance of the leaderboard and code package. All co-authors contribute to maintaining the code package, and Mingyi is responsible for maintaining the leaderboard.

\end{itemize}

\subsection{Details of Baseline Models}
\label{sec::appendix_baseline}


\textbf{VLM2Vec}~\citep{vlm2vec} converts vision-language models (VLMs) into the embedding models capable of handling diverse tasks. It reformulates all tasks as instruction-following ranking problems. Using contrastive learning and task-specific instructions, VLM2VEC learns to produce fixed-dimensional embeddings aligned across modalities. 


\textbf{ColPali}~\citep{colpali} leverages a vision-language model trained to generate high-quality multi-vector embeddings from document page images. Combined with a late interaction matching mechanism, it achieves strong performance on visual document retrieval tasks.

\textbf{LamRA}~\citep{lamra} explores the use of large multimodal models (LMMs) for retrieval, unifying diverse retrieval tasks under a single framework without task-specific fine-tuning. It achieves this by employing two-stage training—language-only pretraining followed by multimodal instruction tuning—to enhance retrieval effectiveness. 

\textbf{GME}~\citep{gme} is a unified multimodal embedding model finetuned from Qwen2-VL. It supports retrieval across single-modal, cross-modal, and fused-modal settings. GME is trained via contrastive learning using a diverse set of multimodal pairs including text, images, and image-text combinations.


\subsection{Details of Benchmark Construction}
\label{sec::appendix_dataset_details}

\subsubsection{Video Retrieval}
\textbf{MSR-VTT}~\citep{msrvtt} is a dataset composed of 10K open-domain videos, each video clip ranging from 10 to 32 seconds in length and accompanied by a total of 200K captions. Following JSFusion~\citep{yu2018joint}, we sampled 1K clip-text pairs to incorporate into our benchmark. The query side contains both the instruction and the video caption, while the candidates consist of all 1K videos.

\textbf{DiDeMo}~\citep{didemo} consists of 10K videos collected from Flickr, each trimmed to a maximum of 30 seconds. Each video includes approximately 3 to 5 annotated pairs of descriptions and their corresponding distinct moments. Following previous work~\citep{liu2019use,luo2021clip4clip}, we concatenate these descriptions and perform ``paragraph-to-video'' retrieval on this benchmark. The official test split, which contains 1,004 paragraph-video pairs, is used.

\textbf{MSVD}~\citep{msvd} contains 80K English descriptions for 1,970 YouTube videos, each ranging from 1 to 62 seconds in length. Each video is annotated with approximately 40 sentences. We use the official test split, which includes 670 videos, and select one sentence per video to construct 670 test cases.

\textbf{YouCook2}~\citep{youcook2} consists of 14K video clips sourced from 2K instructional cooking videos on YouTube. Each video contains multiple actions performed by the chef, accompanied by corresponding textual descriptions and temporal annotations. Each video clip is extracted and annotated with a single sentence. We follow the common practice~\citep{howto100m} of using the validation split and removing videos that also appear in HowTo100M. Different papers may report slightly varying numbers of test cases, typically ranging from 3.1K to 3.3K. Our benchmark includes 3,179 clip-text pairs from YouCook2.

\textbf{VATEX}~\citep{vatex} contains 41,250 video clips sourced from Kinetics-600 dataset and 825K sentence-level descriptions. The public test set originally contained 6K videos. However, since many of them have been removed or set to private and are no longer accessible online, we use only a subset of 4,468 available videos. For each video, we select one description to include in our benchmark.

\subsubsection{Moment Retrieval}
\textbf{QVHighlights}~\citep{qvhighlights} is a dataset comprising 10K videos collected from YouTube, covering a diverse range of topics. Each video is annotated with high-quality labels for both query-based video moment retrieval and highlight detection. In our embedding benchmark, we adopt the standard practice of ranking candidate clips and evaluating performance using Recall@1. In contrast, the QVHighlights paper and some other Vision-Language Models like InternVideo2~\citep{internvideo2} evaluate models using Recall@0.5 and Recall@0.7 with Intersection over Union (IoU) as a threshold, a metric that is not well-suited for embedding-based approaches.

\textbf{Charades-STA}~\citep{charadessta}, derived from the Charades~\citep{sigurdsson2016hollywood} dataset, includes sentence-level temporal annotations for approximately 10K videos. Unlike its predecessor, Charades, Charades-STA replaces annotated action types with temporal sentences that describe actions. To minimize ambiguity in candidate clips, we created a filtered subset of the Charades-STA test set by applying a condition that selects videos where the relevant segment occupies less than one-third of the total video length.

\textbf{MomentSeeker}~\citep{momentseeker} is a dataset designed to benchmark multimodal retrievers on long video moment retrieval tasks. Containing 1.8K queries, MomentSeeker consists of 4 subtasks with various query-side modalities. Additionally, MomentSeeker spans a diverse range of topics, including egocentric videos, cartoons, sports, and movies. For each query, we uniformly sampled nine negative clips and included all the ground truth clips as positive examples.

\subsubsection{Video Classification}
\textbf{Kinetics-700-2020}~\citep{kinetics700} is made up of approximately 648K Youtube video clips, covering a wide range of human actions, around 700 labels in total, such as cooking, driving, and drawing. Each video clip lasts 3 seconds on average. We sampled 1K video answer pairs from the validation set into our benchmark. The candidate texts are the list of all the labels. The raw video data are retrieved from \href{https://github.com/cvdfoundation/kinetics-dataset?tab=readme-ov-file}{CVD Foundation Github}.

\textbf{Something Something v2}~\citep{ssv2} is the updated version of the Something Something v1 dataset. It consists of 220K crowd-source videos focusing on the physical interactions between humans and objects, with an average length of 4.03 seconds and a total of 174 action classes. We randomly sampled 1000 videos-text pairs from the validation split into our benchmark. The candidate texts are the list of all action classes. The raw video data are retrieved from \href{https://www.qualcomm.com/developer/software/something-something-v-2-dataset/downloads}{Qualcomm}. 

\textbf{HMDB51}~\citep{hmdb} is composed of 6K video clips, including both movies and web videos, with 51 action labels, such as catch, drink, and kick. We sampled 1K frames-text pairs from the test splits into our benchmark. The candidate texts are all labels. The raw video data are retrieved from \href{https://serre-lab.clps.brown.edu/resource/hmdb-a-large-human-motion-database/#overview}{the official website}. 

\textbf{Breakfast}~\citep{breakfast} contains around 1.9K crowdsource video clips in the wild, more than 70 hours of total length, which are about preparing for 10 different types of breakfast, such as cereal, milk, pancakes, and fried eggs. There are 6 different camera viewpoints, and we only selected the clips filmed with camera 01, ignoring those filmed by other cameras. We used all the video clips of camera 01, around 433 samples in total. The candidate texts are the 10 types of breakfast. The raw video data are retrieved from
\href{https://serre-lab.clps.brown.edu/resource/breakfast-actions-dataset/#Downloads}{the official website}. 

\textbf{UCF101}~\citep{ucf101} is an open domain video data set consisting of approximately 13K videos with 101 action categories, such as applying makeup, sports and playing instruments. We sampled 1K clip-text pairs from test splits into our benchmarks, and the candidate texts are all the action categories. The raw video data are retrieved from
\href{https://www.crcv.ucf.edu/data/UCF101.php#Results_on_UCF101}{the official website}.

\subsubsection{Video QA}
While embedding models are not primarily designed for open-ended visual question answering, QA tasks offer a valuable way to assess whether a model can effectively understand visual inputs for different downstream purposes. They also enable fair comparison between embedding-based and generation-based approaches. To this end, we select multi-choice QA benchmarks that span a wide range of task types and are relatively less dependent on knowledge or reasoning abilities. We retain the original dataset configurations to ensure compatibility with prior work and facilitate direct comparison with existing models.

\textbf{MVBench}~\citep{mvbench} is a comprehensive benchmark designed to evaluate multi-modal large language models on video understanding, with a particular focus on temporal understanding. It defines 20 video tasks covering a wide spectrum of temporal abilities -- from perception to cognition -- by transforming static tasks into dynamic, multiple-choice QA formats. 

\textbf{Video-MME}~\citep{videomme} is a full-spectrum benchmark for evaluating Multi-modal Large Language Models (MLLMs) on video understanding. It consists of 2,700 manually annotated QA pairs based on 900 videos (totaling 254 hours). It ensures broad scenario coverage and captures diverse temporal dynamics by including a variety of video types and durations.

\textbf{NExT-QA}~\citep{nextqa} is a video question answering benchmark focused on causal and temporal reasoning in untrimmed daily activity videos. It supports both multiple-choice (what we use) and open-ended QA formats. In our study, we utilize only the multiple-choice portion to evaluate models' ability to reason about complex action dynamics and object interactions.

\textbf{EgoSchema}~\citep{egoschema} is a diagnostic benchmark for long-form video understanding, constructed from Ego4D and comprising over 5,000 multiple-choice QA pairs spanning more than 250 hours of egocentric video. Each question is grounded in a 3-minute clip and targets long-range temporal reasoning. In our study, we use a subset of 500 questions for which answer annotations are publicly available.

\subsubsection{Visual Document Retrieval}
\textbf{ViDoRe}~\citep{colpali, vidore_v2} is a benchmark designed to evaluate document retrieval systems. The first version (v1) includes 10 subtasks. The dataset originates from two sources: (1) for academic tasks, it repurposes widely used visual question-answering (VQA) benchmarks, treating each question as a query and the corresponding page as the gold document; (2) for practical tasks, publicly accessible PDF documents are collected, and queries relevant to document pages are generated using Claude-3 Sonnet. 

To address the saturation of the original ViDoRe benchmark, ViDoRe-v2 introduces more realistic and challenging retrieval tasks, including four new diverse and multilingual datasets.

\textbf{VisRAG}~\citep{visrag} serves as the test set for the VisRAG pipeline, which assesses multimodal retrievers in document retrieval. This benchmark consists of six subtasks adapted from VQA datasets, with a filtering process applied to exclude context-dependent queries unsuitable for retrieval.

\textbf{ViDoSeek}~\citep{vidorag} is a large-scale document collection question-answering dataset originally designed to evaluate retrieval-augmented generation (RAG) performance requiring complex reasoning. We adapt it for retrieval by using questions as queries and reference pages as gold images, with each query linked to relevant images from a collection of approximately 5,000 images. The dataset covers diverse content types, including text, charts, tables, and structured layouts.

\textbf{MMLongBench-Doc}~\citep{mmlongbench} is a long-context, multimodal VQA benchmark containing 1,082 expert-annotated questions. Unlike previous VQA datasets, it is built on 135 lengthy PDF documents, averaging 47.5 pages each. To ensure comprehensive evaluation, questions require evidence from multiple sources (text, images, charts, tables, and layout structures) and locations (e.g., specific page numbers). We repurpose this dataset for retrieval, treating questions as queries and evidence pages as gold images, with each query linked to relevant images from a collection of approximately 6,000 images.





\subsection{Detailed Scores}
\label{sec::full_score_table}
\input{tables/full_score_table}

%% file: tables/full_score_table.tex
\definecolor{avgcolor}{RGB}{235,245,255}     
\definecolor{icatcolor}{RGB}{255,240,220}    
\definecolor{vcatcolor}{RGB}{235,255,235}    
\definecolor{vdcatcolor}{RGB}{245,230,255}   

\begin{table}[!ht]
    \centering
    \renewcommand{\arraystretch}{1.1}
    \begin{adjustbox}{width=\textwidth}
    \begin{tabular}{l|cccccccc}
        \toprule
        ~ & ColPali v1.3 & GME-2B & GME-7B & LamRA-Qwen2 & LamRA-Qwen2.5 & VLM2Vec-2B & VLM2Vec-7B & VLM2Vec-V2.0 \\ 
        \midrule
        \rowcolor{avgcolor}
        Avg - All (78 tasks) & 44.4 & 54.1 & 57.8 & 40.4 & 47.4 & 47.0 & 52.3 & 58.0 \\ 
        \midrule
        \rowcolor{avgcolor}
        Avg - Image (36 tasks, Hit@1) & 34.9 & 51.9 & 56.0 & 54.1 & 52.4 & 59.7 & 65.5 & 64.9 \\ 
        \rowcolor{avgcolor}
        Avg - Video (18 tasks, Hit@1) & 28.2 & 33.6 & 38.4 & 35.0 & 33.6 & 28.6 & 33.7 & 34.6 \\ 
        \rowcolor{avgcolor}
        Avg - Visdoc (24 tasks, NDCG@5) & 71.0 & 72.7 & 75.2 & 23.9 & 50.2 & 41.6 & 46.4 & 65.4 \\
        \midrule
        \rowcolor{icatcolor}
        I-CLS (10) & 40.3 & 54.4 & 57.7 & 59.2 & 51.7 & 58.7 & 62.7 & 62.9 \\ 
        \rowcolor{icatcolor}
        I-QA (10) & 11.5 & 29.9 & 34.7 & 26.5 & 34.1 & 49.3 & 56.9 & 56.3 \\ 
        \rowcolor{icatcolor}
        I-RET (12) & 48.1 & 66.9 & 71.2 & 70.0 & 66.9 & 65.0 & 69.4 & 69.5 \\ 
        \rowcolor{icatcolor}
        I-VG (4) & 40.3 & 55.5 & 59.3 & 62.7 & 56.7 & 72.9 & 82.2 & 77.3 \\ 
        \rowcolor{vcatcolor}
        V-CLS (5) & 26.7 & 34.9 & 37.4 & 39.3 & 32.9 & 33.4 & 39.1 & 39.3 \\ 
        \rowcolor{vcatcolor}
        V-QA (5) & 37.8 & 42.0 & 50.4 & 42.6 & 42.6 & 30.5 & 30.0 & 34.3 \\ 
        \rowcolor{vcatcolor}
        V-RET (5) & 21.6 & 25.6 & 28.4 & 24.3 & 23.2 & 20.6 & 29.0 & 28.8 \\ 
        \rowcolor{vcatcolor}
        V-MR (3) & 25.5 & 31.1 & 37.0 & 32.8 & 37.2 & 30.7 & 38.9 & 36.8 \\ 
        \rowcolor{vdcatcolor}
        VD-Vidore-V1 (10) & 83.6 & 86.1 & 89.4 & 22.0 & 56.3 & 49.8 & 56.9 & 75.5 \\ 
        \rowcolor{vdcatcolor}
        VD-Vidore-V2 (4) & 52.0 & 54.0 & 55.6 & 11.5 & 33.3 & 13.5 & 9.4 & 44.9 \\ 
        \rowcolor{vdcatcolor}
        VD-VisRAG (6) & 81.1 & 82.5 & 85.0 & 37.4 & 58.2 & 51.8 & 59.1 & 79.4 \\ 
        \rowcolor{vdcatcolor}
        VD-OOD (4) & 43.1 & 43.1 & 44.4 & 21.0 & 40.1 & 33.5 & 38.1 & 39.4 \\ 
        \midrule
        
        ImageNet-1K & 42.4 & 58.3 & 64.6 & 72.3 & 58.9 & 77.5 & 80.1 & 80.8 \\ 
        N24News & 25.5 & 50.1 & 50.5 & 51.3 & 29.8 & 73.7 & 79.7 & 72.9 \\ 
        HatefulMemes & 50.6 & 52.5 & 53.6 & 49.0 & 51.3 & 58.3 & 69.7 & 56.3 \\ 
        VOC2007 & 69.8 & 75.9 & 80.3 & 80.1 & 78.7 & 74.3 & 80.7 & 85.0 \\ 
        SUN397 & 56.1 & 67.3 & 69.5 & 68.5 & 66.5 & 73.8 & 77.4 & 71.0 \\ 
        Place365 & 27.5 & 35.8 & 39.1 & 40.6 & 37.4 & 35.3 & 37.4 & 35.9 \\ 
        ImageNet-A & 14.9 & 28.8 & 41.2 & 47.0 & 36.3 & 50.9 & 58.1 & 47.4 \\ 
        ImageNet-R & 64.6 & 78.6 & 83.9 & 88.5 & 77.0 & 84.7 & 73.9 & 89.3 \\ 
        ObjectNet & 45.6 & 70.6 & 69.0 & 66.4 & 59.4 & 37.1 & 40.1 & 65.2 \\ 
        Country211 & 6.0 & 26.5 & 24.8 & 28.3 & 21.7 & 21.5 & 29.8 & 25.2 \\ 
        OK-VQA & 9.4 & 29.9 & 33.2 & 37.8 & 39.9 & 48.5 & 56.8 & 51.5 \\ 
        A-OKVQA & 6.6 & 18.6 & 21.0 & 27.0 & 34.1 & 39.5 & 47.3 & 43.6 \\ 
        DocVQA & 11.3 & 29.8 & 41.4 & 22.3 & 37.1 & 82.5 & 89.7 & 90.1 \\ 
        InfographicsVQA & 5.0 & 11.6 & 20.3 & 16.5 & 23.7 & 47.7 & 60.0 & 58.8 \\ 
        ChartQA & 5.7 & 13.4 & 17.8 & 11.7 & 15.0 & 42.3 & 56.9 & 47.4 \\ 
        Visual7W & 6.1 & 16.2 & 22.2 & 19.6 & 24.6 & 51.2 & 52.7 & 52.9 \\ 
        ScienceQA & 16.3 & 27.3 & 28.0 & 26.3 & 31.3 & 30.7 & 38.5 & 38.2 \\ 
        VizWiz & 27.6 & 37.0 & 39.0 & 32.0 & 32.0 & 38.6 & 39.9 & 43.3 \\ 
        GQA & 8.3 & 75.1 & 76.9 & 38.5 & 57.4 & 48.3 & 55.1 & 64.9 \\ 
        TextVQA & 18.8 & 39.7 & 46.8 & 33.0 & 46.1 & 63.3 & 71.6 & 72.2 \\ 
        VisDial & 41.2 & 48.1 & 60.8 & 61.3 & 62.5 & 74.3 & 81.9 & 82.7 \\ 
        CIRR & 8.2 & 44.2 & 54.9 & 51.7 & 44.7 & 46.8 & 51.1 & 57.5 \\ 
        VisualNews\_t2i & 50.1 & 74.7 & 79.7 & 70.4 & 70.1 & 73.1 & 80.5 & 74.5 \\ 
        VisualNews\_i2t & 47.6 & 78.3 & 83.6 & 83.9 & 74.2 & 73.7 & 81.2 & 78.2 \\ 
        MSCOCO\_t2i & 59.2 & 68.1 & 71.2 & 72.2 & 65.7 & 73.4 & 77.2 & 75.3 \\ 
        MSCOCO\_i2t & 49.9 & 63.1 & 57.7 & 73.7 & 71.1 & 68.5 & 73.9 & 71.4 \\ 
        NIGHTS & 65.5 & 67.0 & 67.6 & 65.6 & 64.4 & 66.3 & 67.6 & 68.6 \\ 
        WebQA & 53.8 & 88.8 & 91.4 & 81.0 & 85.7 & 85.9 & 88.3 & 90.6 \\ 
        FashionIQ & 5.9 & 32.9 & 37.8 & 42.0 & 33.4 & 14.0 & 17.1 & 19.5 \\ 
        Wiki-SS-NQ & 80.5 & 73.9 & 78.2 & 69.7 & 67.0 & 54.2 & 62.3 & 66.9 \\ 
        OVEN & 50.0 & 72.3 & 75.1 & 82.0 & 84.8 & 68.3 & 66.5 & 64.3 \\ 
        EDIS & 64.7 & 91.8 & 96.0 & 85.9 & 78.7 & 81.2 & 85.7 & 84.1 \\ 
        MSCOCO & 36.7 & 28.6 & 31.4 & 44.8 & 36.0 & 66.5 & 75.7 & 67.1 \\ 
        RefCOCO & 64.5 & 55.9 & 60.9 & 62.8 & 57.1 & 80.9 & 87.6 & 87.1 \\ 
        RefCOCO-Matching & 3.9 & 73.3 & 78.4 & 75.7 & 82.6 & 75.7 & 84.6 & 85.8 \\ 
        Visual7W-Pointing & 56.1 & 64.1 & 66.5 & 67.3 & 51.2 & 68.3 & 81.0 & 69.2 \\ 
        \midrule
        
        K700 & 23.4 & 35.2 & 39.7 & 42.3 & 32.1 & 31.4 & 35.5 & 38.0 \\ 
        SmthSmthV2 & 25.1 & 29.9 & 30.6 & 36.3 & 25.3 & 30.9 & 32.1 & 42.8 \\ 
        HMDB51 & 24.8 & 43.4 & 47.9 & 40.5 & 33.8 & 33.8 & 42.2 & 40.9 \\ 
        UCF101 & 49.4 & 52.4 & 54.7 & 60.4 & 53.0 & 57.5 & 61.8 & 60.0 \\ 
        Breakfast & 10.9 & 13.6 & 14.3 & 16.9 & 20.1 & 13.4 & 23.8 & 14.8 \\ 
        MVBench & 33.7 & 37.5 & 46.6 & 37.2 & 37.6 & 30.5 & 28.5 & 33.7 \\ 
        Video-MME & 30.6 & 34.3 & 39.2 & 34.1 & 35.1 & 26.9 & 27.8 & 30.7 \\ 
        NExTQA & 35.2 & 39.5 & 53.6 & 43.7 & 44.9 & 20.3 & 20.3 & 20.9 \\ 
        EgoSchema & 38.4 & 40.8 & 46.8 & 44.8 & 47.0 & 25.4 & 21.8 & 34.0 \\ 
        ActivityNetQA & 51.3 & 58.0 & 65.6 & 53.2 & 48.5 & 49.6 & 51.4 & 52.3 \\ 
        DiDeMo & 22.8 & 22.0 & 26.4 & 24.8 & 22.8 & 19.4 & 29.3 & 30.4 \\ 
        MSR-VTT & 17.6 & 27.3 & 31.8 & 22.1 & 25.0 & 25.2 & 34.5 & 28.3 \\ 
        MSVD & 45.4 & 47.6 & 49.7 & 46.1 & 41.9 & 38.2 & 46.7 & 48.1 \\ 
        VATEX & 16.7 & 23.0 & 24.9 & 19.1 & 18.7 & 16.2 & 25.5 & 26.5 \\ 
        YouCook2 & 5.3 & 7.9 & 9.1 & 9.2 & 7.5 & 4.1 & 9.0 & 10.6 \\ 
        QVHighlight & 19.9 & 43.6 & 59.5 & 53.8 & 60.9 & 44.2 & 57.7 & 49.4 \\ 
        Charades-STA & 29.0 & 14.9 & 14.0 & 10.9 & 18.8 & 13.6 & 19.8 & 20.2 \\ 
        MomentSeeker & 27.6 & 34.8 & 37.4 & 33.8 & 31.8 & 34.4 & 39.3 & 40.8 \\ 
        \midrule
        
        ViDoRe\_arxivqa & 81.7 & 82.8 & 86.9 & 10.8 & 53.0 & 48.9 & 60.2 & 80.6 \\ 
        ViDoRe\_docvqa & 56.6 & 53.1 & 57.5 & 19.1 & 25.4 & 27.0 & 34.7 & 44.9 \\ 
        ViDoRe\_infovqa & 84.9 & 90.2 & 91.6 & 46.3 & 72.3 & 67.2 & 70.4 & 83.7 \\ 
        ViDoRe\_tabfquad & 86.9 & 93.3 & 94.6 & 42.8 & 66.1 & 62.6 & 78.2 & 89.2 \\ 
        ViDoRe\_tatdqa & 70.9 & 69.9 & 74.1 & 11.4 & 25.9 & 19.8 & 27.6 & 43.8 \\ 
        ViDoRe\_shiftproject & 75.1 & 89.5 & 96.8 & 12.0 & 27.3 & 41.8 & 38.6 & 60.8 \\ 
        ViDoRe\_artificial\_intelligence & 95.7 & 97.5 & 99.6 & 10.3 & 72.0 & 55.0 & 67.7 & 88.5 \\ 
        ViDoRe\_energy & 94.7 & 91.9 & 95.3 & 24.8 & 65.2 & 59.1 & 60.4 & 86.5 \\ 
        ViDoRe\_government\_reports & 93.6 & 94.6 & 98.8 & 16.4 & 72.2 & 57.1 & 61.8 & 85.0 \\ 
        ViDoRe\_healthcare\_industry & 95.9 & 98.7 & 99.3 & 25.9 & 83.8 & 59.6 & 69.9 & 92.2 \\ 
        ViDoRe\_esg\_reports\_human\_labeled\_v2 & 51.3 & 61.0 & 63.4 & 7.6 & 33.0 & 12.6 & 6.8 & 45.6 \\ 
        ViDoRe\_biomedical\_lectures\_v2\_multilingual & 54.7 & 54.0 & 49.5 & 13.3 & 35.9 & 7.4 & 5.1 & 44.3 \\ 
        ViDoRe\_economics\_reports\_v2\_multilingual & 49.0 & 50.2 & 54.2 & 19.1 & 31.9 & 13.9 & 13.9 & 43.0 \\ 
        ViDoRe\_esg\_reports\_v2\_multilingual & 52.9 & 50.7 & 55.4 & 5.9 & 32.5 & 20.1 & 11.9 & 46.6 \\ 
        VisRAG\_ArxivQA & 80.9 & 82.0 & 87.4 & 2.0 & 37.7 & 41.8 & 52.6 & 76.9 \\ 
        VisRAG\_ChartQA & 78.2 & 79.9 & 81.9 & 41.3 & 65.9 & 57.9 & 70.2 & 84.4 \\ 
        VisRAG\_MP-DocVQA & 86.8 & 84.4 & 89.2 & 33.4 & 54.5 & 43.2 & 52.8 & 71.8 \\ 
        VisRAG\_SlideVQA & 95.0 & 93.4 & 94.5 & 56.5 & 76.5 & 74.0 & 72.8 & 91.5 \\ 
        VisRAG\_InfoVQA & 85.7 & 91.4 & 93.5 & 56.3 & 73.3 & 70.7 & 72.0 & 85.7 \\ 
        VisRAG\_PlotQA & 60.3 & 64.1 & 63.4 & 34.6 & 41.2 & 23.4 & 34.4 & 66.1 \\ 
        ViDoSeek-page & 22.2 & 21.6 & 23.2 & 11.3 & 23.1 & 17.7 & 22.3 & 21.9 \\ 
        ViDoSeek-doc & 83.7 & 83.6 & 83.9 & 37.1 & 80.3 & 74.3 & 77.8 & 80.2 \\ 
        MMLongBench-page & 14.2 & 15.8 & 16.2 & 8.0 & 13.5 & 9.6 & 11.8 & 11.9 \\ 
        MMLongBench-doc & 52.5 & 51.4 & 54.3 & 27.6 & 43.5 & 32.6 & 40.5 & 43.7 \\ 
    \bottomrule
    \end{tabular}
    \end{adjustbox}
\caption{
Performance comparison of various models on the full MMEB-v2 benchmark, covering 78 tasks across image, video, and visual document modalities. Numbers in parentheses represent the task count for each category.}
\label{tab:full_score}
\end{table}